\def\year{2021}\relax
\documentclass[letterpaper]{article} 
\usepackage{aaai21}  
\usepackage{times}  
\usepackage{helvet} 
\usepackage{courier}  
\usepackage[hyphens]{url}  
\usepackage{graphicx} 
\usepackage{graphicx}  
\usepackage{natbib}  
\usepackage{caption} 
\usepackage{amsmath}
\usepackage{multirow}
\usepackage{multicol}
\usepackage{array, boldline, makecell, booktabs}
\usepackage{amssymb}
\usepackage{color}
\newcommand{\tabincell}[2]{\begin{tabular}{@{}#1@{}}#2\end{tabular}}
\usepackage{bbm}
\usepackage[switch]{lineno}
\usepackage{bm}

\newcommand{\tabref}[1]{Table~\ref{#1}}
\newcommand{\figref}[1]{Figure~\ref{#1}}

\newcommand{\eg}[1]{\textit{e.g.,}}
\newcommand{\ie}[1]{\textit{i.e.,}}

\title{Group-Wise Semantic Mining for Weakly Supervised Semantic Segmentation}

\author{Xueyi Li\textsuperscript{\rm 1}, 
		Tianfei Zhou\textsuperscript{\rm 2}\thanks{Corresponding author: \textit{Tianfei Zhou}}, 
		Jianwu Li\textsuperscript{\rm 1},
		Yi Zhou\textsuperscript{\rm 3},
		Zhaoxiang Zhang\textsuperscript{\rm 4} \\
		}
 
\affiliations{
 	\textsuperscript{\rm 1 }{Beijing Key Laboratory of Intelligent Information Technology,} \\ {School of Computer Science and Technology, Beijing Institute of Technology, China} \\
 	\textsuperscript{\rm 2 }{Computer Vision Laboratory, ETH Zurich, Switzerland} \\
 	\textsuperscript{\rm 3 }{School of Computer Science and Engineering, Southeast University, China} \\
 	\textsuperscript{\rm 4 }{Center for Research on Intelligent Perception and Computing, CASIA, China}\\
 	\texttt{\{xueyili,ljw\}@bit.edu.cn~~~ tianfei.zhou@vision.ee.ethz.ch}
 }

\begin{document}

\maketitle

\begin{abstract}
	
	Acquiring sufficient ground-truth supervision to train deep visual models has been a bottleneck over the years due to the data-hungry nature of deep learning. This is exacerbated in some structured prediction tasks, such as semantic segmentation, which requires pixel-level annotations. This work addresses weakly supervised semantic segmentation (WSSS), with the goal of bridging the gap between image-level annotations and pixel-level segmentation. We formulate WSSS as a novel group-wise learning task that explicitly models semantic dependencies in a group of images to estimate more reliable pseudo ground-truths, which can be used for training more accurate segmentation models. In particular, we devise a graph neural network (GNN) for group-wise semantic mining, wherein input images are represented as graph nodes, and the underlying relations between a pair of images are characterized by an efficient co-attention mechanism. Moreover, in order to prevent the model from paying excessive attention to common semantics only, we further propose a graph dropout layer, encouraging the model to learn more accurate and complete object responses. The whole network is end-to-end trainable by iterative message passing, which propagates interaction cues over the images to progressively improve the performance. We conduct experiments on the popular PASCAL VOC 2012 and COCO benchmarks, and our model yields state-of-the-art performance. Our code is available at: \textit{\texttt{https://github.com/Lixy1997/Group-WSSS}}.
\end{abstract}
\section{Introduction}

\begin{figure*}[!t]
	\centering
	\includegraphics[width=\textwidth]{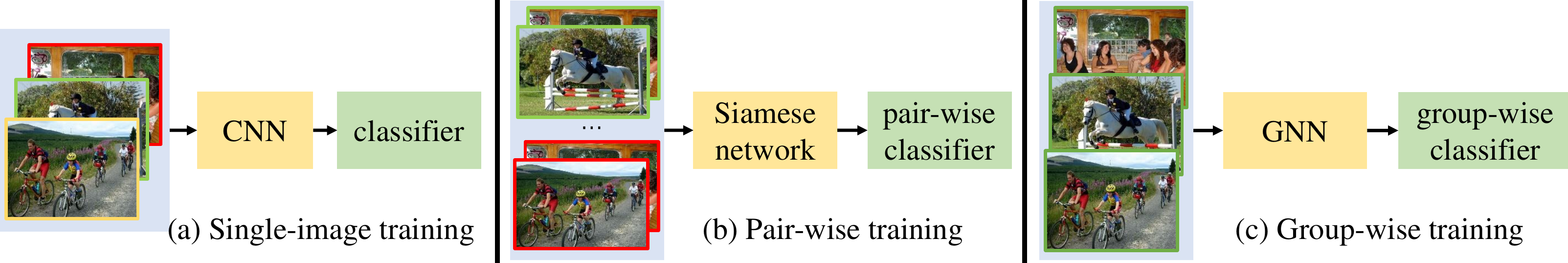}
	\caption{\small\textbf{Architecture comparison of existing frameworks \textit{vs.} Ours.} (a) Single-image models feed each image one by one into the network for training, which bears high similarity with standard classifiers (\eg, VGG). (b) Pair-wise methods extract features from a pair of images using a Siamese network, and make predictions using a pair-wise classifier which has learned the correlation between the two images. (c) We propose a group-wise method that accepts an arbitrary number of images as input. The input images are \textit{iteratively} processed by a GNN to enable substantial information to exchange, and a group-wise classifier is then adopted for prediction.}
	\label{fig:1}
\end{figure*}

Semantic segmentation is a fundamental task in computer vision, aiming to assign a semantic category to each pixel in an image. It can benefit a wide variety of applications including autonomous driving, image editing and medical diagnosis.  With the recent renaissance of deep neural networks, semantic segmentation has achieved tremendous progress. However, most of the leading approaches~\cite{long2015fully,wang2019learning,zhou2020matnet,zhou2020motion} are fully supervised, requiring massive amounts of pixel-level annotated training data, which are extremely expensive and time-consuming to obtain. In contrast, the weak supervision alternatives, \eg, image-level tags~\cite{pathak2015constrained,kolesnikov2016seed,qi2016augmented,wei2016stc,chaudhry2017discovering,ahn2018learning,fan2018associating}, scribbles~\cite{lin2016scribblesup,vernaza2017learning} or bounding-box annotations~\cite{dai2015boxsup,khoreva2017simple,song2019box}, are less costly. Thus, it is of interest to explore the potential of these weak supervision cues in providing a data-efficient solution for semantic segmentation. In this paper, we aim to address weakly supervised semantic segmentation (WSSS) under the supervision of image-level tags, which can be obtained effortlessly.

WSSS based on image tags is extremely challenging because fine-grained pixel-level annotations, which are typically required for semantic segmentation, are difficult to obtain from class labels. The pioneering work,~\cite{zhou2016learning}, proposes an efficient and straightforward way to solve this by recognizing the discriminative regions specific to a given category using class activation maps (CAMs), which are then refined to obtain pseudo ground-truths for supervising a semantic segmentation network. Along this line, a number of approaches have been proposed to improve the estimation of CAMs so that they cover the full extent of objects rather than only the most representative parts. For example, some approaches~\cite{wei2017object,kolesnikov2016seed,choe2019attention} manipulate internal feature maps to guide the network to perceive easily ignored but essential parts, while others~\cite{hou2018self,chang2020weakly,fan2020learning,wang2020self} adopt self-ensembling or self-supervision to improve localization.

However, the mainstream methods above are merely based on \textit{\textbf{single images}} (\figref{fig:1} (a)), ignoring the valuable semantic context existing in a group of images. The very recent studies~\cite{fan2020cian,sun2020mining} utilized Siamese networks to model the relations between a pair of images, leading to a \textit{\textbf{pair-wise}} solution (\figref{fig:1} (b)). These approaches have proven effective in locating more accurate object regions. However, seeking relations between two images at a time is still limited in capturing substantial semantic context. 
Accordingly, we introduce a more promising, and fundamentally different  \textit{\textbf{group-wise}} solution (\figref{fig:1} (c)) which comprehensively mines richer semantics from a group of images. Our main motivation is that the availability of group images containing instances of the \textit{same semantic classes} can make up for the absence of detailed supervisory information. From this perspective, we hypothesize that it is desirable to take advantage of all available information for WSSS, including not only individual image properties, but also group-level synergetic relationships.

Based on the above analysis, we propose a novel deep learning model for WSSS. Unlike previous pair-wise approaches, our model is targeted at group-wise semantic mining to capture more comprehensive relations among input images. Specifically, we develop an efficient, end-to-end trainable graph neural network (GNN), and conduct recursive reasoning for group-wise semantic understanding. In our graph, the nodes represent a group of input images, and edges describe pair-wise relations between two connected images.  We consider two images as connected only if they share common semantic objects with each other, and their relation is then characterized by an elaborately designed co-attention mechanism. Through iterative message passing, the information from individual elements can be efficiently integrated and broadcasted over the graph structure. In this way, our model is capable of leveraging explicit semantic dependencies among images to obtain better node representations. However, this graph reasoning strategy mainly focuses on co-occurring semantics in a group of images, ignoring isolated objects. To address this, we further introduce a graph dropout layer, which can be seamlessly integrated into the GNN for iterative inference. The graph dropout layer selectively suppresses the most salient objects, forcing the network to be biased toward their counterparts. 

Our method has two appealing characteristics over single-image and pair-wise methods. \textbf{First}, it is capable of learning semantic relations from an arbitrary number of images using a flexible GNN framework. The GNN empowers our model to inherit the complementary strengths of neural networks in learning capability and graphical models in structure representations. \textbf{Second}, our model adopts multi-step, iterative inference to progressively improve image representations. This is more favorable than directly producing representations by one-step inference in previous approaches

In summary, our main contributions are three-fold: \textbf{First}, we demonstrate the advantages of group-wise semantic mining for WSSS, and proffer a graph-aware solution for effective inference. \textbf{Second}, we develop a graph dropout layer to promote the missing categories, leading to more accurate localization. \textbf{Third}, we evaluate the proposed approach on two large-scale benchmarks, \ie, PASCAL VOC 2012~\cite{everingham2010pascal} and COCO~\cite{lin2014microsoft}, and the results demonstrate its superiority.

\section{Related Work}

\noindent\textbf{Weakly Supervised Semantic Segmentation.}
Recent years have seen	a surge of interest in semantic segmentation under weak supervision (\eg, image-level labels, scribbles, bounding boxes), greatly reducing human efforts in manual labeling. In particular, methods operating with image-level labels have attracted the most attention since they require minimal annotation efforts. Most of these methods follow a popular pipeline that trains an image classifier using image-level labels, and exploits CAMs to highlight the most discriminative object regions for a particular semantic category as its pseudo ground-truth. However, CAMs are weak in revealing complete object regions, resulting in poor segmentation performance. Some pioneering efforts address this difficulty 
by learning pixel affinities~\cite{ahn2018learning}, erasing the most discriminative parts~\cite{wei2017object,choe2019attention,lee2019ficklenet}, optimizing intra-class discrimination~\cite{fan2020learning}, or applying region growing~\cite{kolesnikov2016seed,wei2018revisiting,huang2018weakly} to capture the full extent of objects. However, these methods are confined to using only limited image-level information. More recent approaches thus follow the self-supervised paradigm to acquire additional supervisions~\cite{shimoda2019self,wang2020self}, or rely on Siamese networks to capture semantic relations between a pair of images~\cite{fan2020cian,sun2020mining}. 

In this paper, we take a further step toward discovering higher-order relations among images. A graph model is designed to encode such relationships. Through graph reasoning, our model iteratively refines object representations by accepting informative knowledge from other images.

\begin{figure*}[!t]
	\centering
	\includegraphics[width=\textwidth]{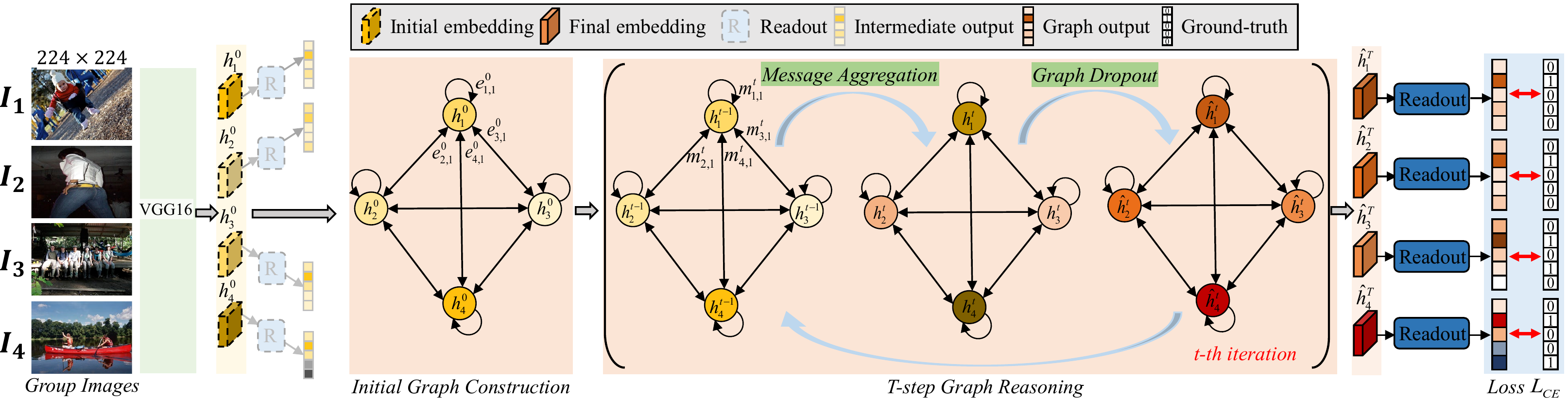}
	\put(-403,17){\tiny $\bm{l}_4^m$}
	\put(-403,40){\tiny $\bm{l}_3^m$}
	\put(-403,91){\tiny $\bm{l}_2^m$}
	\put(-403,111){\tiny $\bm{l}_1^m$}
	\put(-31,106){\tiny $\bm{l}_1^g$}
	\put(-31,79){\tiny $\bm{l}_2^g$}
	\put(-31,53){\tiny $\bm{l}_3^g$}
	\put(-31,28){\tiny $\bm{l}_4^g$}
	\put(-14,106){\tiny $\bm{l}_1$}
	\put(-14,79){\tiny $\bm{l}_2$}
	\put(-14,53){\tiny $\bm{l}_3$}
	\put(-14,28){\tiny $\bm{l}_4$}
	\put(-355,12){\tiny $\mathcal{G}=(\mathcal{V}, \mathcal{E})$}
	
	\caption{\small \textbf{Overview of the proposed group-wise semantic mining network} during the training phase. Given a group of images (\ie, $\{\bm{I}_i\}_{i=1}^4$), our model uses VGG16 to extract convolutional features (\ie, $\bm{h}_i^0\}_{i=1}^4$), which are used as the initial embeddings for graph construction. Next, our model conducts $T$-step graph reasoning to iteratively refine the features by \textit{message passing} (Eq.\!~\eqref{eq:8}), \textit{message aggregation} (Eq.\!~\eqref{eq:2}), and \textit{graph dropout} (Eq.\!~\eqref{eq:10}). The final features (\ie, $\{\hat{\bm{h}}{}^T_i\}_{i=1}^4$) are fed into a readout function (Eq.\!~\eqref{eq:9}) to get the  predictions (\ie, $\{\bm{l}_i^g\}_{i=1}^4$).}

	\label{fig:2}
\end{figure*}
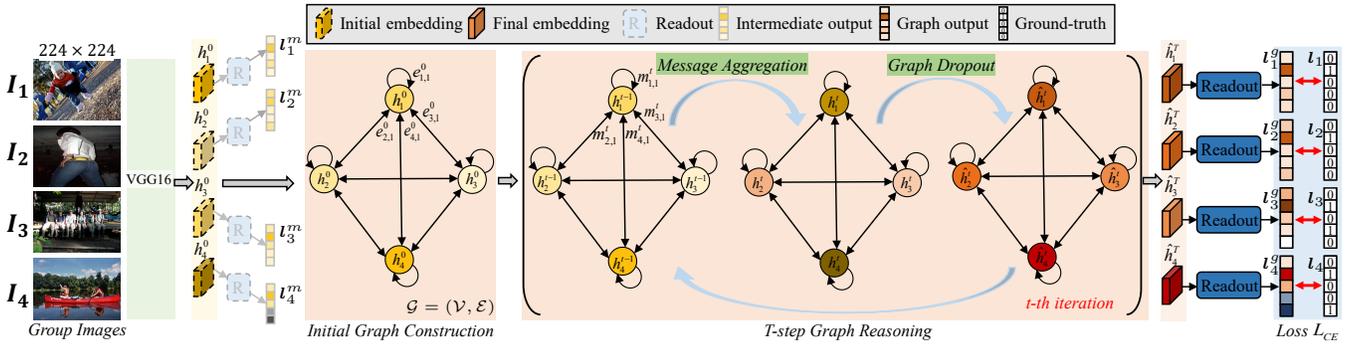

\noindent\textbf{Graph Neural Networks.}
Graph neural networks were proposed in~\cite{scarselli2008graph}, and have since gained widespread attention due to their superiority in dealing with flexible graph-structured data. GNNs typically model the graph elements (\eg, nodes, edges) and approximation inference as learnable neural networks, and conduct iterative reasoning to explicitly discover the relations among nodes. They have achieved wide success in a variety of fields, including molecular biology~\cite{gilmer2017neural}, computer vision~\cite{qi20173d,lu2020video,marino2016more,wang2019zero,santoro2017simple,wang2020hierarchical}, and machine learning~\cite{velivckovic2017graph,qu2019gmnn}. Inspired by these efforts, we build an image-level graph network to model their semantic relations for the WSSS task. Assisted by a graph dropout layer, our model can generate more accurate pseudo ground-truths for semantic segmentation.

\section{Methodology}

In this section, we elaborate on the proposed model for WSSS. Given training images with only image-level labels, current efforts operate on two sub-tasks to achieve pixel-wise predictions. The first one is \textit{pseudo ground-truth generation}, which relies on an image classification network to localize discriminative regions. The other one is \textit{semantic segmentation}, which conducts dense predictions using a fully convolutional network (FCN) under the supervision of pseudo labels. Our approach also follows this pipeline. However, unlike previous approaches that treat each single image independently, our model aims to mine common semantic patterns from multiple images by graph inference. In this way, our model can alleviate the incomplete-annotation problem in WSSS and produce more accurate pseudo labels.

%

\subsection{Preliminary: Graph Neural Networks}
We start by revisiting the basic concept of GNNs. We define a graph $\mathcal{G}\!=\!(\mathcal{V}, \mathcal{E})$ by its node set $\mathcal{V}\!=\!\{v_1,\ldots,v_n\}$ and edge set $\mathcal{E}\!=\!\{e_{i,j}\!=\!(v_i, v_j)|v_i,v_j\in\mathcal{V}\}$. We assume that each node $v_i$ is associated with a feature embedding vector $\bm{h}_i$, and each edge $e_{i,j}$ has an edge representation $\bm{e}_{i,j}$. During inference, GNNs iteratively improve the feature representations at a node by aggregating its neighborhood features. Specifically, a GNN maps the graph $\mathcal{G}$ to the node outputs through two phases: a message passing phase and a readout phase. The message passing phase is defined in terms of a \textit{message function} $\mathcal{F}_M$, whose input is a node's features and output is a message, and an \textit{aggregation function} $\mathcal{F}_A$, whose input is a set of messages and output is the updated features. Suppose we conduct $T$ rounds of message passing; the $t$-th round for a node $v_i$ can be described as:
\begin{align}
 &\text{message passing:} ~~~  \bm{m}_i^t = \sum_{v_j\!\in\!\mathcal{N}_i} \mathcal{F}_M^t (\bm{h}_i^{t-1}, \bm{h}_j^{t-1}, \bm{e}_{i,j})\label{eq:1}, \\
&\text{message aggregation:} ~~~ \bm{h}_i^t = \mathcal{F}_A(\bm{h}_i^{t-1},  \bm{m}_i^t)\label{eq:2},
\end{align}
where for $v_i$, the message function firstly summarizes the information (\ie, $\bm{m}_i^t$) from its neighbors $\mathcal{N}_i$, and then uses it to update the node state. Then, in the readout phase, a task-specific readout function $\mathcal{F}_R$ operates on the final node representation $\bm{h}_i^T$ to produce a node output:
\begin{equation}\small
\text{readout phase:} \quad o_i = \mathcal{F}_R(\bm{h}_i^T).
\end{equation}

Next, we will present the details of the proposed graph-based semantic mining model for pseudo ground-truth generation in WSSS.

\subsection{Group-Wise Semantic Mining Network}

\noindent\textbf{Problem Definition:} 
Given a collection of training samples, our first goal is to generate corresponding pseudo ground-truths, which will later be used to supervise semantic segmentation networks. To achieve this, we formulate the problem as graph-based semantic co-mining among multiple images. Formally, we denote $\mathcal{I}\!=\!\{(\bm{I}_i, \bm{l}_i)\}_{i=1}^N$ as the training data, where $\bm{I}_i\!\in\!\mathbb{R}^{w\!\times\!h\!\times\!3}$ is an image and $\bm{l}_i\!\in\!\{0,1\}^L$ is the corresponding image-level ground-truth with $L$ possible semantic categories. During training, we \textit{selectively} sample $K$ images $\{\bm{I}_i\}_{i=1}^K$ as a mini-batch, and model their relations as a directed graph $\mathcal{G}\!=\!(\mathcal{V},\mathcal{E})$, where the image $\bm{I}_i$ is denoted as node $v_i\!\in\!\mathcal{V}$, and the relation between $v_i$ and $v_j$ is represented by edge $e_{i,j}\!\in\!\mathcal{E}$. To better capture more comprehensive common semantics, we consider two nodes $v_i$ and $v_j$ to be linked only if there is at least one semantic category shared between them. Besides, we assume that every node has a \textit{self-edge}, \eg, $e_{i,i}$ for $v_i$.

Given the above definitions, our network aims to conduct pseudo ground-truth generation in a graph learning scheme, under the full supervision of image-level labels as well as the implicit semantic relations among different images. In this manner, our model can capture richer semantic information and obtain more accurate pseudo labels. Next, we describe the details of each component in our model.

\noindent\textbf{Node Embedding:}
As an initial step, we abstract a high-level feature representation for each input image. Formally, given $\bm{I}_i$, we extract features $\bm{h}_i\!\in\!\mathbb{R}^{W\!\times\!H\!\times\!C}$ from the convolutional stages of a standard classification network (\eg, VGG\!~\cite{simonyan2014very}). The embedding of node $v_i$ is then initialized by $\bm{h}_i$, which is a $(W,H,C)$-dimensional tensor preserving full spatial details for more effective pixel-level matching during graph reasoning. 

\noindent\textbf{Edge Embedding:} For each edge $e_{i,j}$ connecting $v_i$ to $v_j$, we aim to learn an edge embedding $\bm{e}_{i,j}^t$ at each iteration $t$ to find the correct semantic correspondence between the two nodes. This is achieved by dense matching over node embeddings using the following bilinear model:
\begin{equation}\small\label{eq:4}
\bm{e}_{i,j}^t = \bm{h}_i^t\bm{W}\bm{h}_j^{t\top} \in \mathbb{R}^{WH\!\times\!WH},
\end{equation}
where $\bm{h}_i^t\!\in\!\mathbb{R}^{WH\!\times\!C}$ and $\bm{h}_j^t\!\in\!\mathbb{R}^{WH\!\times\!C}$ are flattened into matrix representations for computational convenience. $\bm{W}\in\mathbb{R}^{C\!\times\!C}$ is a trainable weight matrix. In Eq.~\eqref{eq:4}, $\bm{e}_{i,j}^t$ encodes the similarity between $\bm{h}_i^t$ and $\bm{h}_j^t$ for all pairs of spatial locations.
For the edge $e_{j,i}$, its embedding at iteration $t$ is simply calculated as $\bm{e}_{j,i}^{t} = \bm{e}_{i,j}^{t\top}$.

It should be noted that Eq.~\eqref{eq:4} introduces a large number of parameters, increasing the computational cost. To alleviate this, $\bm{W}$ is approximately factorized into two low-rank matrices $\bm{P}\!\in\!\mathbb{R}^{C\!\times\!\frac{C}{d}}$ and $\bm{Q}\!\in\!\mathbb{R}^{C\!\times\!\frac{C}{d}}$, where $d~(d\!>\!1)$ is a reduction ratio. Then, Eq.~\eqref{eq:4} can be rewritten as:
\begin{equation}\small\label{eq:5}
\bm{e}_{i,j}^t = \bm{h}_i^t\bm{P}\bm{Q}^\top\!\bm{h}_j^{t\top} \in \mathbb{R}^{WH\!\times\!WH}.
\end{equation}
Eq.~\eqref{eq:5} has significant advantages over Eq.~\eqref{eq:4} in both model parameters and computational cost: 1) it reduces the number of parameters by $2/d$ times; 2) it only requires $(2WHC^2\!+\!W^2H^2C)/d$ multiplication operations, instead of the $WHC^2+W^2H^2C$ in Eq.~\eqref{eq:4}. 

In addition, for each self-edge $e_{i,i}$, its embedding $\bm{e}_{i,i}$ captures the self-relation over the node representation $\bm{h}_i$. We compute $\bm{e}_{i,i}^t$ at iteration $t$ by self-attention~\cite{vaswani2017attention,wang2018non}, which can effectively capture long-range semantic dependencies. In particular, the self-attention calculates the response at a position by attending to all the positions within the same node embedding:
\begin{equation}\small\label{eq:6}
\bm{e}_{i,i}^t= \text{softmax}(\phi_f(\bm{h}_i^t)\phi_g^\top(\bm{h}_i^t))\phi_h(\bm{h}_i^t) + \bm{h}_i^t\!\in\!\mathbb{R}^{W\!\times\!H\!\times\!C},
\end{equation}
where $\phi_{\{f,g,h\}}$ are $1\!\times\!1$ convolutional operators. As seen, we also consider it to be a residual layer in Eq.~\eqref{eq:6}, which can effectively preserve information in the original feature map.

\noindent\textbf{Message Passing:}
Given the node and edge embeddings, our model iteratively updates the hidden states of graph nodes by applying message functions to collect information from their neighboring nodes. More specifically, for a node $v_i$, it absorbs knowledge along two types of edges: 1) a self-edge $e_{i,i}$ that encodes rich context-aware knowledge in $v_i$; and 2) other edges $\{e_{j,i}\}_j$ that connect $v_j$ to $v_i$. For the former, our message function directly reads the message from $\bm{e}_{i,i}$, \ie, $\bm{m}_{i,i}^t\!=\!\bm{e}_{i,i}^{t-1}$; while for the latter, the messages are summarized as: 
\begin{equation}\small\label{eq:7}
\bm{m}_{j,i}^t\!=\! \text{softmax}_r(\bm{e}_{i,j}^{t-1})\bm{h}_{j}^{t-1} \!\in\!\mathbb{R}^{WH\!\times\!C},
\end{equation}
where $\text{softmax}_r$ denotes the row-wise softmax operation. In Eq.\!~\eqref{eq:7}, we accumulate knowledge from $\bm{h}_{j}^{t-1}$, which is weighted based on the similarity between $\bm{h}_{i}^{t-1}$ and $\bm{h}_{j}^{t-1}$. $\bm{m}_{j,i}^t$ is then reshaped to a $(W,H,C)$-dimensional tensor. Then, we can easily summarize the message for $v_i$ at  the $t$-th iteration as:
\begin{equation}\small\label{eq:8}
\bm{m}_i^{t} = \sum_{v_j\!\in\!\mathcal{N}_i} \bm{m}_{j,i}^{t-1} + \bm{m}_{i,i}^{t-1}.
\end{equation}

Next, the aggregation function $A$ updates the hidden states of nodes, as given in Eq.\!~\eqref{eq:2}. In our method, $A$ is instantiated by a ConvGRU network~\cite{ballas2015delving}, which is an extension of the GRU update function used in~\cite{gilmer2017neural}. In this way, the message passing algorithm runs for $T$ steps before convergence, iteratively collecting messages and updating node embeddings.

\noindent\textbf{Readout Phase:}
Having repeated the above process for $T$ time steps, we obtain the final node embedding $\bm{h}_i^T\!\in\!\mathbb{R}^{W\!\times\!H\!\times\!C}$ for $v_i$. Then, the readout function $R$ is applied to the features $\bm{h}_i^T$ for image classification:
\begin{equation}\small\label{eq:9}
\bm{l}_i^g = \mathcal{F}_R(\bm{h}_i^T) = \text{GAP}(\phi_r(\bm{h}_i^T)) \in \mathbb{R}^L,
\end{equation}
where $\phi_r$ is a class-aware convolutional layer with kernel size $1\!\times\!1$ that obtains a feature map with $L$ channels, and $\text{GAP}$ denotes a \textit{global average pooling} layer which produces the final classification outputs.


\noindent\textbf{Pseudo Ground-Truth Generation by Self-Ensembling:} 
Once the classification results are obtained (Eq.\!~\eqref{eq:9}), we discover the discriminative image regions for a particular category following~\cite{jiang2019integral}. These regions are further thresholded to generate pseudo ground-truths. 

Besides, as shown in~\figref{fig:2}, for each input image, our network produces two outputs based on raw convolutional features $\bm{h}_i^0$ as well as enriched features $\bm{h}_i^T$. This not only introduces additional deeply supervised constraints~\cite{lee2015deeply} which could benefit the performance, but also enables the results to be further improved by ensembling the CAMs of multiple outputs. We found that the pseudo ground-truths from different outputs are well complementary with each other, and self-ensembling them by averaging can further improve the performance (see~\tabref{table:2}).

\subsection{Graph Dropout Layer}
The above graph reasoning scheme enables our model to discover common semantics present in different images (Eq.\!~\eqref{eq:5}). The features of these semantics can be accordingly enriched by summarizing all the information from other images (Eq.\!~\eqref{eq:8}). However, standalone categories, which may exist only in a single image, are almost ignored in this procedure. To address this, we introduce a graph dropout layer to force the network to pay more attention to these categories. Formally, given the feature map $\bm{h}_i^t\!\in\!\mathbb{R}^{W\!\times\!H\!\times\!C}$ at the $t$-th iteration, we average it along the channel dimension to obtain $\bm{o}_i^t\!\in\!\mathbb{R}^{W\!\times\!H}$. Then, we generate a mask $\bm{s}^t_i\!\in\mathbb{R}^{W\!\times\!H}$ as follows:
\begin{equation}\label{eq:10}
\bm{s}^t_i\!=\!\begin{cases}
\text{sigmoid}(\bm{o}_i^t), ~~~~~~~~~~~~~~~~~~~~~~~\text{if} ~~r < \delta_r;\\
\bm{o}_i^t \mathbbm{1}(\bm{o}_i^t < \text{max}(\bm{o}_i^t) * \delta_d), ~~~\text{otherwise}.
\end{cases}
\end{equation}
Here, the parameter $\delta_r$ is a drop rate threshold, determining whether to carry out the dropout operation or not. The parameter $r$ is a scalar generated from a random generator. If $r\!<\!\delta_r$, $\bm{s}^t_i$ is an importance map which supports the activations in $\bm{h}_i^t$; otherwise, the layer drops the highly activated semantic regions to emphasize standalone semantics. $\mathbbm{1}(x)$ is a matrix indicator function which returns $1$ for the true elements in $x$, and $0$ otherwise. The $\text{max}(\cdot)$ operation calculates the maximum value for a 2D tensor. $\delta_d$ is a threshold controlling the dropout.
Finally, we enhance the feature maps by:
\begin{equation}\label{eq:11}
\hat{\bm{h}}{}^t_i = \bm{h}_i^t \otimes \bm{s}^t_i,
\end{equation}
where $\otimes$ denotes spatial-wise multiplication. Note that $\hat{\bm{h}}{}^t_i$ is then used to replace original features $\bm{h}_i^t$ in the next iteration.

\begin{table}[t]
	\centering
	\small
	\caption{\small \textbf{Quantitative comparison of different methods} on PASCAL VOC 2012 \textit{val} and \textit{test} in terms of mIoU. $^*$: VGG backbone. $^\dagger$: ResNet backbone.}
	\begin{tabular}{l|c|cc}
		\hlineB{2.5}
		\textit{Methods} & \textit{Pub.}  &  \textit{Val} & \textit{Test} \\ \hline\hline

		$^*$MEFF{\!~\tiny~\cite{ge2018multi}} & CVPR18 & - & 55.6\% \\ 
		$^*$GAIN{\!~\tiny~\cite{li2018tell}} & CVPR18 & 55.3\% & 56.8\% \\ 
		$^*$MDC{\!~\tiny~\cite{wei2018revisiting}}  & CVPR18  & 60.4\% & 60.8\% \\
		$^*$RRM{\!~\tiny~\cite{zhang2020reliability}}  & AAAI20  & 60.7\% & 61.0\% \\	
		$^\dagger$MCOF{\!~\tiny~\cite{wang2018weakly}} & CVPR18  & 60.3\% & 61.2\% \\
		$^\dagger$SeeNet{\!~\tiny~\cite{hou2018self}}  & NIPS18  & 63.1\% & 62.8\% \\
		$^\dagger$DSRG{\!~\tiny~\cite{huang2018weakly}} & CVPR18  & 61.4\% & 63.2\% \\
		$^\dagger$AffinityNet{\!~\tiny~\cite{ahn2018learning}} & CVPR18  & 61.7\% & 63.7\% \\
		$^\dagger$SS-WSSS{\!~\tiny~\cite{araslanov2020single}} & CVPR20  & 62.7\% & 64.3\% \\
		$^\dagger$SSNet{\!~\tiny~\cite{zeng2019joint}} & ICCV19  & 63.3\% & 64.3\% \\
		$^\dagger$IRNet{\!~\tiny~\cite{ahn2019weakly}} & CVPR19  & 63.5\% & 64.8\% \\ 
		$^\dagger$CIAN{\!~\tiny~\cite{fan2020cian}}  & AAAI20  & 64.3\% & 65.3\% \\
		$^\dagger$FickleNet{\!~\tiny~\cite{lee2019ficklenet}}  & CVPR19  & 64.9\% & 65.3\% \\
		$^\dagger$IAL{\!~\tiny~\cite{wang2020weakly}} & IJCV20 & 64.3\% & 65.4\% \\
		$^\dagger$SSDD{\!~\tiny~\cite{shimoda2019self}}  & ICCV19  & 64.9\% & 65.5\% \\
		$^\dagger$SEAM{\!~\tiny~\cite{wang2020self}}  & CVPR20  & 64.5\% & 65.7\% \\
		$^\dagger$SubCat{\!~\tiny~\cite{chang2020weakly}}  & CVPR20  & 66.1\% & 65.9\% \\
		$^\dagger$OAA+{\!~\tiny~\cite{jiang2019integral}}  & ICCV19  & 65.2\% & 66.4\% \\
		$^\dagger$RRM{\!~\tiny~\cite{zhang2020reliability}}  & AAAI20  & 66.3\% & 66.5\% \\
		$^\dagger$BES{\!~\tiny~\cite{chenweakly}}  & ECCV20  & 65.7\% & 66.6\% \\
		$^\dagger$EME{\!~\tiny~\cite{fanemploying}}  & ECCV20  & 67.2\% & 66.7\% \\
		$^\dagger$MCIS{\!~\tiny~\cite{sun2020mining}}  & ECCV20  & 66.2\% & 66.9\% \\
		$^\dagger$ICD{\!~\tiny~\cite{fan2020learning}} & CVPR20  & 67.8\% & 68.0\% \\ \hline
		$^*$Ours (VGG16)  & -- & 63.3\% & 63.6\% \\
		$^\dagger$Ours (ResNet101) & -- & \textbf{68.2\%} & \textbf{68.5\%} \\ 
		\hline
	\end{tabular}
	
	\label{table:1}
\end{table}

\subsection{Detailed Network Architecture}

Our model is comprised of two sub-networks: a \textit{classification network} for group-wise pseudo ground-truth generation and a \textit{segmentation network} for semantic segmentation. 

\noindent\textbf{Classification Network.}
We choose VGG16~\cite{simonyan2014very} as the backbone, which is pre-trained on ImageNet~\cite{deng2009imagenet}. We replace the last convolutional layer in VGG16 by dilated convolutions with a rate of 2, and the feature maps from this layer are taken as the initial node representations for the GNN. For each image $\bm{I}_i$, our network has two outputs: an intermediate output $\bm{l}^m_i$ which is directly obtained from the backbone (\figref{fig:2}), and a final output $\bm{l}^g_i$ after graph reasoning (\figref{fig:2}). Then, the loss function of the classification network for image $i$ is:
\begin{equation}\small\label{eq:12}
\mathcal{L} = \mathcal{L}_{\text{CE}}\left(\bm{l}_i^g,\bm{l}_i \right) + \lambda\mathcal{L}_{\text{CE}}\left(\bm{l}_i^m,\bm{l}_i \right),
\end{equation}  
where $\mathcal{L}_{\text{CE}}$ indicates the standard sigmoid cross entropy loss, and $\lambda$ balances the two losses.

After training, we obtain the CAMs for each training image from the two classification layers mentioned earlier, and combine them to obtain foreground object seeds. Besides, we also follow conventional practices~\cite{jiang2019integral,fan2020learning} to estimate background seeds using an off-the-shelf salient object detection model~\cite{hou2017deeply}. The final pseudo labels are generated by combining the foreground and background seeds.

\noindent\textbf{Segmentation Network.}
Following~\cite{chang2020weakly,fan2020cian}, we choose DeepLab-v2~\cite{chen2017deeplab} as the segmentation network due to its superior performance in fully supervised semantic segmentation tasks. 

\begin{table}[t]
	\centering
	\small
	\caption{\small \textbf{Quantitative comparison of different methods} on COCO \textit{val} in terms of mIoU. All methods use VGG16 as the backbone.}
	\begin{tabular}{l|c|cc}
		\hlineB{2.5}
		\textit{Methods} & \textit{Pub.}  &  \textit{Val}\\ \hline\hline
		BFBP{\!~\tiny~\cite{saleh2016built}} & ECCV16 & 20.4\%  \\ 
		SEC{\!~\tiny~\cite{kolesnikov2016seed}} & ECCV16 & 22.4\% \\ 	
		DSRG{\!~\tiny~\cite{huang2018weakly}}  & CVPR18  & 26.0\% \\
		IAL{\!~\tiny~\cite{wang2020weakly}}  &IJCV20  & 27.7\% \\
		\hline
		Ours & -- & \textbf{28.4\%}\\ \hline
	\end{tabular}
	
	\label{table:3}
\end{table}

\section{Experiments}
\subsection{Experimental Setup}

\noindent\textbf{Datasets:}
We conduct our experiments on two datasets: PASCAL VOC 2012~\cite{everingham2010pascal} and COCO~\cite{lin2014microsoft}. 1) \textbf{PASCAL VOC 2012} is currently the most popular benchmark for WSSS. The dataset contains 20 semantic categories (\eg, person, bicycle, cow) and one background category. Following standard protocol~\cite{huang2018weakly,lee2019ficklenet,wang2020self}, extra data from SBD~\cite{hariharan2011semantic} is also used for training, leading to a total of 10,582 training images. We evaluate our model on the standard validation and test sets, which have 1,449 and 1,456 images, respectively. 
2) \textbf{COCO} is a more challenging benchmark with 80 semantic classes. Since more complex contextual relations exist among these categories, it is interesting to examine the performance of our model in this dataset. Following~\cite{wang2020weakly}, we use the default train/val splits (80k images for training and 40k for validation) in the experiment.

\noindent\textbf{Evaluation Metric:} For fair comparison, we utilize a widely used metric~\cite{wang2018weakly,choe2019attention,sun2020mining}, \textit{mean Intersection-over-Union (mIoU)}, for evaluation. The scores on the test set of PASCAL VOC are obtained from the official evaluation server.

\noindent\textbf{Training Details:} 
1) \textit{Greedy Mini-Batch Sampling.} During training, we design a heuristic, greedy strategy to sample $K$ training images in each mini-batch. Starting from a randomly sampled image $\bm{I}_i$, we further find another $K\!-\!1$ images, each of which shares as many common semantic objects with $\bm{I}_i$ as possible. These $K$ images are then used to build a $K$-node GNN. This sampling strategy enables our model to better explore rich relationships among groups of images and improve the results.
2) \textit{Training Settings.} 
For the classification network, the number of nodes $K$ and message passing steps $T$ in the GNN are separately set to 4 and 3 by default. The input image size is $224\!\times\!224$. The entire network is trained using the SGD optimizer with initial learning rates of 1e-3 for the backbone and 1e-2 for the GNN, which are reduced by 0.1 every five epochs. The total number of epochs, momentum and weight decay are set to 15, 0.9, and 5e-4, respectively. The $\lambda$ in Eq.~\eqref{eq:12} is empirically set to 0.4 and the $d$ in Eq.~\eqref{eq:5} is set to 4. For the segmentation network, we follow the training setting in~\cite{chen2017deeplab}, but use the generated pseudo ground-truths as the supervision.

\noindent\textbf{Reproducibility:} 
Our network is implemented in PyTorch and trained on four NVIDIA RTX 2080Ti GPUs with 11GB memory per card. The testing is conducted on the same machine with one GPU card. 

\subsection{Performance on PASCAL VOC 2012}
We evaluate the proposed approach on PASCAL VOC 2012 against current top-performing WSSS methods that only operate with image-level labels. Following conventions, we evaluate the performance of our model using VGG16~\cite{simonyan2014very} and ResNet101~\cite{he2016identity} as the backbones, respectively. As reported in~\tabref{table:1}, our model with ResNet101 achieves the best performance, scoring $\textbf{68.2\%}$ and $\textbf{68.5\%}$ on the \textit{val} and \textit{test} sets, respectively. It significantly outperforms the current leading approach, \ie, ICD~\cite{fan2020learning}, by $\textbf{+0.4\%}$ and $\textbf{+0.5\%}$ on the two evaluation sets. 

In addition,~\tabref{table:1} also shows that the proposed approach outperforms both pair-wise models (\ie, CIAN~\cite{fan2020cian} and MCIS~\cite{sun2020mining}), and all single-image based models (\eg, RRM~\cite{zhang2020reliability}, OAA+~\cite{jiang2019integral}), by a large margin. The reason lies in that existing methods exploit limited context in image collection, while our approach can learn more effective inter-image representations with GNNs.



\begin{figure*}[!t]
	\centering
	\includegraphics[width=0.99\textwidth]{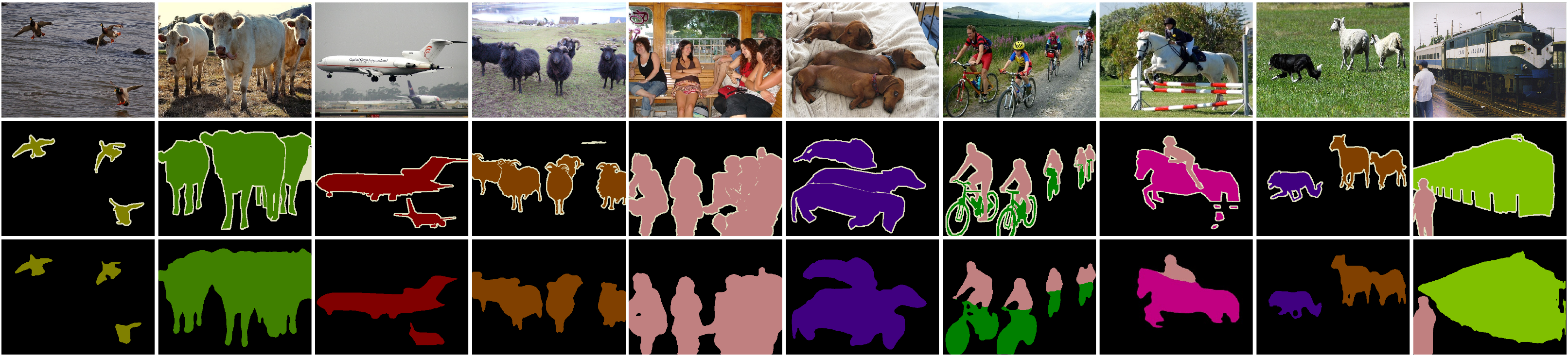}
	\caption{\small\textbf{Qualitative results} on PASCAL VOC 2012 \textit{val}. From top to bottom: input images, ground-truths, and our segmentation results.}
	\label{figure:3}
\end{figure*}

In~\figref{figure:3}, we also provide sample results for representative images in PASCAL VOC 2012 \textit{val}. The images cover various challenging factors in WSSS, such as multiple objects, different semantic categories, small objects, and cluttered background. We see that our model can deal with these difficulties well, resulting in appealing segmentation results.


\subsection{Performance on COCO}
We further examine the performance of our model on COCO. As reported in~\tabref{table:3}, our model achieves the best mIoU score (\ie, \textbf{28.4\%}) on the validation set, outperforming the second-best result, \ie, IAL~\cite{wang2020weakly}, by \textbf{0.7\%}. This further proves the superiority of our model.

\begin{table}[t]
	\centering
	\caption{\small \textbf{Diagnostic experiments of our model} on PASCAL VOC 2012 \textit{val} in terms of mIoU. For all variants, we use ResNet101 as the backbone.}
	\small
	\begin{tabular}{c|c|c|c|c}
		\hlineB{2.5}
		\multicolumn{2}{c|}{\textit{Aspect}} & \multicolumn{2}{c|}{\textit{Variant}} & \textit{mIoU}\\\hline\hline
		
		\multicolumn{2}{c|}{\multirow{2}{*}{\tabincell{c}{\textbf{Full Model}}}} & \multicolumn{2}{c|}{$T=3$, $K=4$} & \multirow{2}{*}{\tabincell{c}{\textbf{68.2}\%} }\\
		
		\multicolumn{2}{c|}{}&\multicolumn{2}{c|}{$\delta_r=0.8$, $\delta_d=0.7$} & \\ \hline
			
		& \multirow{3}{*}{\tabincell{c}{{Node}\\{Number}} } & \multicolumn{2}{c|}{$K=3$} & 68.1\% \\
		& & \multicolumn{2}{c|}{$K=5$} & 67.8\%\\
		& & \multicolumn{2}{c|}{$K=6$} & 67.6\%\\ \cline{2-5}	
		
		\multirow{3}{*}{\tabincell{c}{{Graph}\\{Reasoning}}} 
		& \multirow{3}{*}{\tabincell{c}{{Message}\\{Passing}} } & \multicolumn{2}{c|}{$T=2$} & 67.8\% \\
		& & \multicolumn{2}{c|}{$T=4$} & 68.0\%\\
		& & \multicolumn{2}{c|}{$T=5$} & 68.0\%\\ \cline{2-5}

		& \multirow{5}{*}{\tabincell{c}{{Graph}\\{Dropout}}}  
		& \multirow{2}{*}{\tabincell{c}{$\delta_r=0.8$} } &
		$\delta_d=0.9$ & 68.0\% \\
		& & & $\delta_d=0.5$ & 67.7\%\\  \cline{3-5}
		
		& & $\delta_r=0.6$ & \multirow{2}{*}{\tabincell{c}{$\delta_d=0.7$} } & 66.8\% \\
		& & $\delta_r=0.4$ &  & 63.6\%\\ \cline{3-5}
		
		& & \multicolumn{2}{c|}{\textit{w/o} dropout} & 67.7\% \\ \hline
		
		\multicolumn{2}{c|}{\multirow{3}{*}{\tabincell{c}{Self-Ensembling}}} & \multicolumn{2}{c|}{intermediate output} & 64.1\%\\ 
		\multicolumn{2}{c|}{}&\multicolumn{2}{c|}{graph output} & 67.8\% \\
		\multicolumn{2}{c|}{}&\multicolumn{2}{c|}{self-ensembling} & 68.2\% \\
		
		\hline
	\end{tabular}
	\label{table:2}
\end{table}

\begin{figure}[!t]
	\centering
	\includegraphics[width=0.47\textwidth]{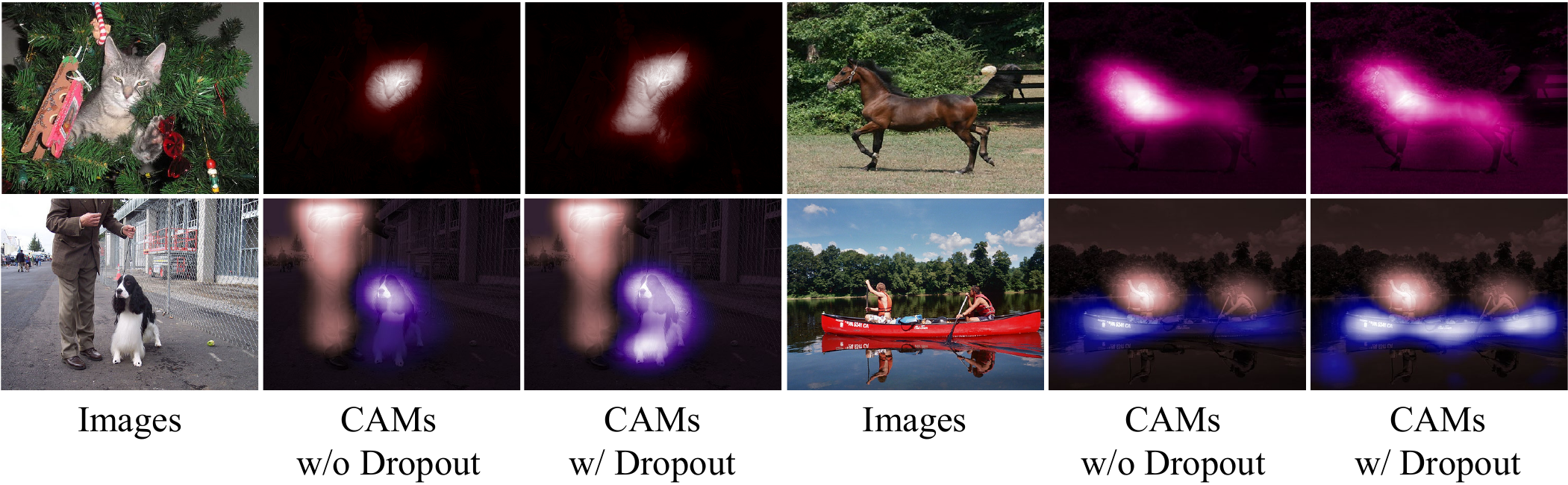}
	\caption{\small\textbf{Visual comparisons of CAMs} generated \textit{w/} or \textit{w/o} the graph dropout layer.}
	\label{fig:4}
\end{figure}

\begin{figure}[!t]
	\centering
	\includegraphics[width=0.47\textwidth]{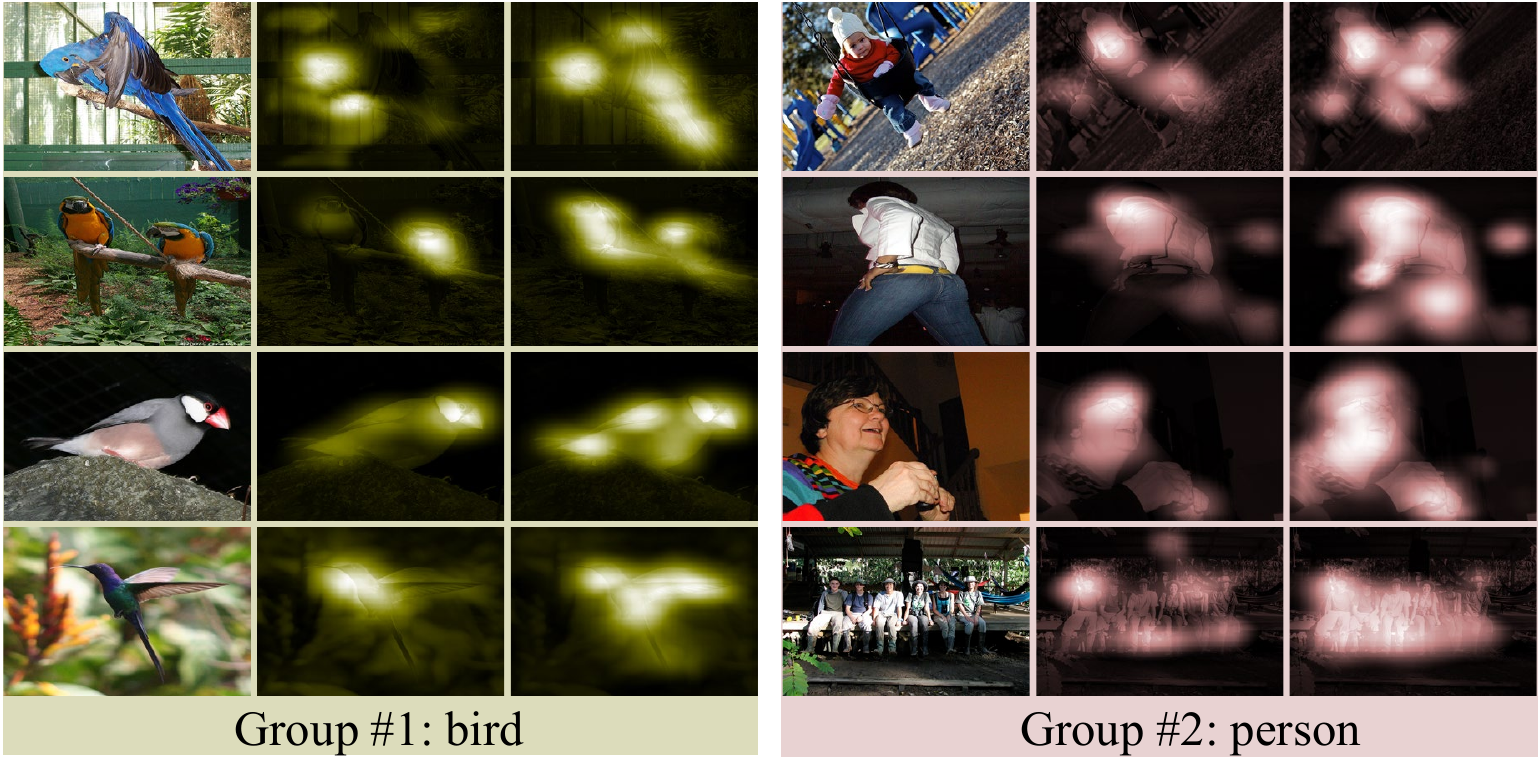}
	
	\caption{\small \textbf{Visual comparisons of CAMs}. Here we provide the results of two groups of images. For each group, we show the input images, CAMs from the \textit{intermediate readout layer} and CAMs from the \textit{graph readout layer} (from left to right). Our model clearly provides more accurate CAMs after group-wise graph reasoning.}
	
	\label{fig:5}
\end{figure}

\subsection{Diagnostic Experiments}
We further conduct diagnostic analysis on PASCAL VOC 2012 \textit{val} set to verify the effectiveness of the essential modules in our approach. We use ResNet101 as the default backbone for all the studies. The performance of our full model with default parameters is given in the first row of~\tabref{table:2}.


\noindent\textbf{Number of Nodes $K$:} We first investigate the effect of the node number $K$ used in the GNN, which indicates the number of images in a group. As shown in~\tabref{table:2}, the model achieves comparably high performance with three or four nodes. However, when more nodes are added, the performance decreases significantly. This can be attributed to the trade-off between meaningful semantic relations and noise brought by group images. When $K\!=\!3$ or $4$, the semantic relations can be fully exploited to improve the integral regions of objects. However, when more images are further considered, meaningful semantic cues reach a bottleneck and noise, introduced by imperfect localization of the classifier, dominates, thus leading to performance degradation.

\noindent\textbf{Number of Message Passing Steps $T$:}
We further evaluate the impact of the message passing steps by comparing the performance with different $T$ ranging from 2 to 5. From~\tabref{table:2}, we observe that the mIoU score is significantly improved when $T$ varies from 2 to 3. The performance decreases slightly when considering more steps. Therefore, we set $T\!=\!3$ as default for message passing.



\noindent\textbf{Graph Dropout Layer:}
To verify the effectiveness of the proposed graph dropout layer, we design multiple experiments to search the optimal configuration of parameters \textit{drop-rate} and \textit{drop-th}. We observe that both parameters have great influences on the performance. As observed in~\tabref{table:2}, our model reaches the best performance at $\delta_r\!=\!0.8$ and $\delta_d\!=\!0.7$. If $\delta_d$ is higher (\eg, 0.9), most discriminative regions will be kept, and thus ignored regions will remain unactivated. In contrast, if the $\delta_d$ is lower, the regions with high responses  will be excessively dropped, leading to degraded classification accuracy. 

In addition, the parameter $\delta_r$ controls whether to drop the responses or not during training. As shown in~\tabref{table:2}, a $\delta_r$ of $0.8$ helps to achieve the best mIoU score. Such a setting not only maintains the classification ability of the network by keeping discriminative regions with a high probability, but also drives the network to mildly attend to other regions. We can also see that by setting $\delta_r$ to smaller values (\eg, $0.6$ or $0.4$), the performance encounters a significant decrease.

Moreover, we examine the performance of our model without the graph dropout layer. As seen, without the dropout layer, the performance of our model decreases by $0.5\%$ in terms of mIoU, which reveals its importance.

Finally, we illustrate some examples of the final CAMs generated \textit{with} or \textit{without} the graph dropout layer. As shown in~\figref{fig:4}, without the dropout layer, the network only focuses on the most discriminative parts (\eg, heads of the cat and the horse). This is improved with our dropout layer, which helps to highlight non-discriminative object regions.

\noindent\textbf{Self-Ensembling:}
In addition to the supervision on the final outputs, we also introduce deep supervision signals on the intermediate features. Such multi-level supervision has proven effective for improving the performance of various vision tasks. Besides, this enables us to combine the multiple outputs with low cost to further boost the performance. Here, we examine the self-ensembling strategy by building three network variants, \ie, \textit{intermediate output}, \textit{graph output} and \textit{self-ensembling}, in which the final CAMs are separately extracted from the intermediate readout layer, graph-aware readout layer, and their ensemble, respectively. As shown in~\tabref{table:2}, the \textit{intermediate output} only obtains an mIoU score of $64.1\%$, greatly lagging behind the $67.8\%$ obtained by the \textit{graph output}. This demonstrates that through iterative graph reasoning, our model can improve the image representations by integrating information from group images, leading to huge performance gains. Furthermore, the self-ensembling strategy boosts the performance to $68.2\%$.

In~\figref{fig:5}, we illustrate two groups of images with their CAMs from the \textit{intermediate readout layer} and \textit{graph readout layer}. As seen, in both groups, the CAMs are well-refined to cover more complete foreground regions after graph reasoning. Besides, in many cases, the CAMs from two output layers complement with each other well, enabling better results to be obtained by self-ensembling.

\section{Conclusion}

In this paper, we have introduced a group-wise learning framework for weakly supervised semantic segmentation (WSSS). We formulate the task within a graph neural network (GNN), which operates on a group of images and explores their semantic relations for representation learning. By iterative graph reasoning, our model provides better pseudo ground-truths, which further lead to significant performance improvement for the semantic segmentation results. We also devise a graph dropout layer to facilitate the discovery of complete object regions.
We conduct extensive experiments on PASCAL VOC 2012 and COCO benchmarks, and the results demonstrate that the proposed approach performs favorably against the state-of-the-art methods.


\bibliographystyle{aaai}
\small\bibliography{ref}

\end{document}